\documentclass[letterpaper, 10 pt, conference]{ieeeconf}  
\IEEEoverridecommandlockouts                              
    
\usepackage{amsmath,amssymb,amsfonts}
\usepackage{algorithmic}
\usepackage{graphicx}
\usepackage{textcomp}
\usepackage{xcolor}
\usepackage{soul}
\usepackage{array}
\usepackage{tabularx}
\usepackage{svg}
\usepackage{subcaption}
\usepackage{makecell}
\usepackage{multirow}
\usepackage{hyperref}
\usepackage{nicefrac}
\usepackage{acronym}

\usepackage{amsmath}
\usepackage{mathtools}

\acrodef{ML}{Machine Learning}
\acrodef{DL}{Deep Learning}
\acrodef{DNN}{Deep Neural Network}
\acrodef{UAV}{Unmanned Aerial Vehicle}
\acrodef{RQ}{Research Question}
\acrodef{mAP}{Mean Average Precision}
\acrodef{AP}{Average Precision}
\acrodef{mAR}{Mean Average Recall}
\acrodef{AR}{Average Recall}
\acrodef{DTD}{Describable Textures Dataset}
\acrodef{NPS}{Non-printability Score}
\acrodef{TV}{Total Variation}
\acrodef{IOU}{Intersection Over Union}
\acrodef{ASR}{Attack Success Ratio}

\title{\LARGE \bf
Model Agnostic Defense against Adversarial Patch Attacks on Object Detection in Unmanned Aerial Vehicles
}

\author{Saurabh Pathak$^{1}$, Samridha Shrestha$^{1}$ and Abdelrahman AlMahmoud$^{1}$
\thanks{$^{1}$The authors are  with Secure Systems Research Center (SSRC) at Technology Innovation Institute (TII), Abu Dhabi.
{\tt\small \{saurabh, samridha, abdelrahman\}@ssrc.tii.ae}}%
}

\begin{document}

\maketitle
\thispagestyle{empty}
\pagestyle{empty}

\begin{abstract}
Object detection forms a key component in \acp{UAV} for completing high-level tasks that depend on the awareness of objects on the ground from an aerial perspective. In that scenario, adversarial patch attacks on an onboard object detector can severely impair the performance of upstream tasks. This paper proposes a novel model-agnostic defense mechanism against the threat of adversarial patch attacks in the context of \ac{UAV}-based object detection. We formulate adversarial patch defense as an occlusion removal task. The proposed defense method can neutralize adversarial patches located on objects of interest, without exposure to adversarial patches during training. Our lightweight single-stage defense approach allows us to maintain a model-agnostic nature, that once deployed does not require to be updated in response to changes in the object detection pipeline. The evaluations in digital and physical domains show the feasibility of our method for deployment in \ac{UAV} object detection pipelines, by significantly decreasing the \acl{ASR} without incurring significant processing costs. As a result, the proposed defense solution can improve the reliability of object detection for \acp{UAV}.
\end{abstract}

\section{Introduction}
\label{sec:intro}
The utilization of \acp{UAV} has seen a substantial surge in recent times. A report suggests that the market for these vehicles is projected to quadruple by the year 2030~\cite{precedenceresearch}. \acp{UAV} find applications in a variety of areas, including cargo transportation, remote sensing, and surveillance operations. Object detection is a crucial component in the automation of these vehicles~\cite{Ebrahimi2021}. The object detection module, typically powered by \acp{DNN}, constantly analyzes the camera feed from the \ac{UAV}. The automation systems of the \acp{UAV} heavily depend on the accuracy and reliability of the \ac{DNN} object detector for high-level tasks such as tracking a detected object and effectively communicating it to an operator~\cite{Franke2022}. Consequently, it is imperative to ensure that the object detectors in \ac{UAV} deployments are reliable and robust against potential threats.

\begin{figure}[t!]
    \centering
    \begin{subfigure}[b]{0.49\columnwidth}   
        \centering 
        \includegraphics[width=.55\textwidth]{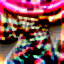}
        \caption{Adversarial patch example}
    \end{subfigure}
    \begin{subfigure}[b]{0.49\columnwidth}   
        \centering 
        \includegraphics[width=\textwidth]{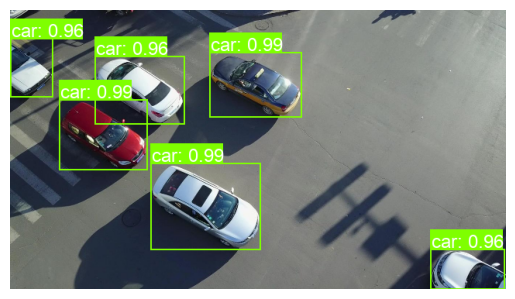}
        \caption{Before attack}
    \end{subfigure}
    \begin{subfigure}[b]{0.49\columnwidth}   
        \centering
        \includegraphics[width=\textwidth]{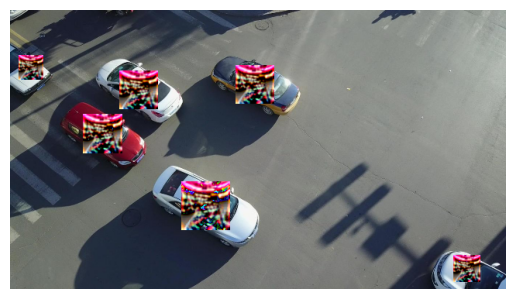}
        \caption{After attack}
        \label{subfig:postatkexample}
    \end{subfigure}
    \begin{subfigure}[b]{0.49\columnwidth}   
        \centering
        \includegraphics[width=\textwidth]{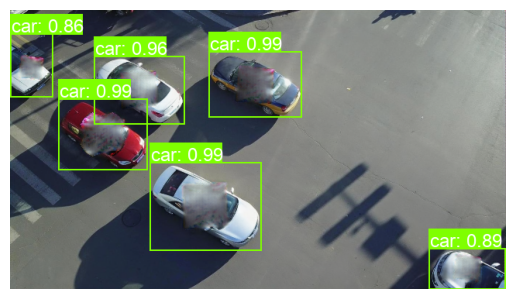}
        \caption{After defense}
        \label{subfig:postdefexample}
    \end{subfigure}
    \caption{Example of an adversarial patch being used to conceal vehicles from an EfficientDet-lite$4$ model on \textit{VisDrone} dataset.}
    \label{fig:prob_statement_examplepatch}
\end{figure}
In recent years, it has been shown that \acp{DNN} are vulnerable to evasion attacks, which are triggered by the addition of specifically crafted input perturbations~\cite{Yin_2022_WACV}. In the context of object detection, a common method of attack focuses on adding adversarial patches directly to the object of interest. An advantage of this type of attack is that an adversary with no access to the camera feed can focus on specific types of objects in an image and transfer the patches to the physical domain by printing the patches and attaching them to actual objects that are then imaged by the camera~\cite{Thys_2019_CVPR_Workshops}.
In the physical domain, adversarial patches serve as a form of ``camouflage''~(Fig.~\ref{subfig:postatkexample}) that can disrupt the functionality of \acp{DNN}-based object detectors, impairing their capacity to accurately identify objects. Given their aerial perspective, \acp{UAV} often need to survey extensive physical areas within a single image frame. This scenario exacerbates the issue, as the objects under consideration by a \ac{UAV} appear smaller relative to the image size due to the altitude and viewpoint. Consequently, a patch-based attack on object detection can have devastating impacts on a higher-level task such as surveillance, tracking, or delivery. Despite this, a significant number of existing object detectors developed for \ac{UAV} applications do not take into account the potential presence of adversaries, operating under the assumption that their systems will not be subjected to this threat.~\cite{MITTAL2020104046}.

Adversarial patches have demonstrated their effectiveness in misleading object detectors for tasks such as causing misclassification of traffic signs~\cite{WeiTPAMI}, making people invisible to detection systems~\cite{Thys_2019_CVPR_Workshops}, or even obscuring cars~\cite{du2022physical}.
However, adversarial patch generation against the \ac{UAV}-based object detectors is still mostly overlooked in the literature~\cite{Klingneriros}.
Specifically, a patch generation process for \ac{UAV}-based aerial imagery must take into account changes in the camera viewing angle and perspective and must be effective at a wide range of distances, significantly affecting the generation challenge~\cite{Chawlairos}.
Despite reports demonstrating the feasibility of adversarial patches against satellite imagery~\cite{du2022physical}, there has been minimal focus in existing literature examining the degree to which adversarial patches can undermine the reliability of object detectors for \acp{UAV}.

In addition to the threat of adversarial patches to \acp{UAV}, current defense strategies in the literature often involve the use of adversarial training methods, which may include the incorporation of adversarial patches during the training phase. While adversarial training can significantly mitigate the effects of adversarial attacks, it often comes at the expense of model accuracy. Furthermore, the increasing reliance on third-party solutions has led to the common use of open-sourced or pretrained models from external sources. These models, often frozen and quantized for deployment on resource-limited platforms, may not have been trained with a specific focus on adversarial threats. Consequently, additional defense mechanisms are needed to protect these models after deployment. This is often achieved by adding a preprocessing stage to counteract the effects of adversarial patches~\cite{xiang2021patchguard, Liu_2022_CVPR, chiang2021adversarial}. However, this solution necessitates the use of adversarial patches that are specific to the downstream model, while also requiring updates to the defense solution on the \ac{UAV} to account for new patch generation techniques~\cite{shafahi2020universal}. Additionally, if the model is replaced, the solution must also be retrained.

\textbf{Contributions.} 
This paper introduces a novel model-agnostic defense mechanism against adversarial patches for object detection in \acp{UAV}. This mechanism is implemented as a preprocessing stage, where the defense against adversarial patches is formulated as an occlusion removal task executed through a thin convolutional autoencoder. Consequently, our method does not rely on any model-specific assumptions about the generation of the adversarial patch, making it robust to changes in the patch generation process. Decoupling from the patch generation process allows our method the flexibility to work with various downstream object detection models without the need for modification or retraining. Experimental results demonstrate that our approach can effectively neutralize adversarial patches prior to the object detection task independently of the downstream model.

In summary, our paper's main contributions are:

\begin{itemize}
    \item An assessment of the impact of adversarial patches on common \ac{UAV} object detectors.
    Our experiments reveal that adversarial patches can significantly affect object detection performance in \acp{UAV}, achieving up to $84$\% attack success rate.
    \item A novel model-agnostic defense mechanism to facilitate reliable object detection in the presence of adversarial patches.
    Our defense approach can reduce the \ac{ASR} of adversarial patch attacks by $\approx 30$\% on average, and can also reduce the impact of non-adversarial patch occlusions while registering an average per-image additional processing cost of only $\approx 4$\% on the detection pipeline.
\end{itemize}

The remainder of this paper is organized as follows. 
Section~\ref{sec:background} further describes the object detection reliability on \acp{UAV} and presents related works.
Section~\ref{sec:problemstatement} evaluates how adversarial patches affect \acp{UAV} object detectors.
Section~\ref{sec:proposal} describes the proposed model-agnostic defense model.
Section~\ref{sec:evaluation} evaluates the proposed scheme and Section~\ref{sec:conclusion} concludes our work.

\section{Background}
\label{sec:background}
\subsection{UAV Object Detection}
Object detection in \acp{UAV} typically follows a four-phase process~\cite{MITTAL2020104046}. Initially, the \ac{UAV} camera feed is continuously collected through a \textit{Data Acquisition} module. Subsequently, the ingested camera images undergo preprocessing before being utilized for object detection. This may involve encoding, resizing, and normalizing the ingested image appropriately. The processed image is then sent to an \textit{Object Detection} module, which employs a Deep Neural Network (\ac{DNN}) to identify predetermined object classes. These detected objects can be used for autonomous decision-making on the \ac{UAV}, such as reporting to the operator or tracking the object with the \ac{UAV}. Given the resource-limited nature of onboard electronics, it is generally preferable to maintain a lightweight object detection model and pipeline. This approach helps ensure low latency and power consumption, which are critical factors for the efficient operation of \acp{UAV}.

\subsection{Adversarial Patches in UAV Applications}
An adversarial patch is designed to alter the pixels of an input image before it is processed by the object detector of an \ac{UAV}~\cite{du2022physical}. To accomplish this, an adversary undertakes an optimization process to create a patch that, when superimposed on a specific object, causes the target object detector to misidentify it. More precisely, given an object detection function $f(x): x \in X \rightarrow y \in Y$ that outputs an object label $y$ based on the input image $x$, the goal of the adversarial attack is to find a patch $x^*$ that, when placed over the object $x$, biases the target detector toward an incorrect prediction.

An adversarial patch begins as a fixed-size arrangement of random noise. This is scaled to a specific proportion of the target object's bounding box size and positioned on each target object. The modified image is then inputted into the target object detection model. The backpropagated gradients from the object detector are utilized to update the pixel values of the patch. The optimization process is typically designed to minimize the classification confidence score of the target object detector with respect to the target objects~\cite{Thys_2019_CVPR_Workshops}.
Fig. \ref{fig:prob_statement_examplepatch} shows an example of an adversarial patch that can impact vehicle identification.

\subsection{Related Works} \label{subsec:related}
Over the last years, several works have shown that adversarial patches can significantly affect the accuracy of state-of-the-art \ac{DNN}-based object detectors~\cite{MITTAL2020104046}.
In general, proposed techniques are evaluated in the digital domain, where the generated patch is digitally overlaid on the targeted object.
T. B. Brown \emph{et al.}~\cite{brown2017adversarial} proposed one of the first approaches to generate adversarial patches.
Their scheme generates patches to be digitally placed on a given image to bias the classifier.
Their work significantly decreases the classifier accuracy, however, it does not address the printability aspects of the generated patch.
K. Eykholt \emph{et al.}~\cite{eykholt2018robust} proposed a patch-based attack that generates stickers to be printed on traffic signs.
They showed that it significantly affects the reliability of object detection schemes in the physical domain.
Since then, a plethora of works have shown the efficacy of adversarial patches against a wide range of applications, from traffic sign detection~\cite{WeiTPAMI}, obscuring people~\cite{Thys_2019_CVPR_Workshops}, or even cars~\cite{du2022physical}.
Yet their threat to \ac{UAV}-related applications is still in its beginnings.
Andrew Du \emph{et al.}~\cite{du2022physical} aimed the adversarial patch generation against cars on aerial imagery.
The authors were able to affect the detection accuracy in the physical domain significantly.
Unfortunately, they used a satellite image dataset that does not account for the challenges related to the \ac{UAV} domain, in particular, the high variance in camera viewing angles, distances, and perspectives.
Similarly, J. Lian \emph{et al.}~\cite{LianTGRS} showed the effectiveness of adversarial patches on satellite imagery to conceal airplanes. However, adversarial patch impact on \ac{UAV} object detection remains largely unexplored, with some recent exceptions~\cite{inproceedings,rs14215298} studying the YOLO variants in the context.

As a result of the adversarial patch threat to \ac{DNN}-based object detection, defense techniques are also the subject of several works in the literature~\cite{xiang2021patchguard}.
In general, proposed schemes are implemented by adding a preprocessing stage~\cite{xiang2021patchguard,Liu_2022_CVPR,chiang2021adversarial}, modifying the \ac{DNN} architecture~\cite{shafahi2020universal}, or retraining the model with the adversarial patches included~\cite{Raoeccv}.
To this extent, current solutions usually do not consider their solution's processing costs.
In practice, \acp{UAV} are resource-constrained devices that should execute their tasks with minimal processing footprint while maintaining their reliability.
Consequently, as \ac{UAV} applications are usually not considered in the adversarial patch literature, proposed solutions cannot be easily used for their defense.
In such a case, deployed \ac{DNN}-based object detectors are not easily updated and must execute their tasks with low processing needs while being resilient to adversaries.

\section{Problem Statement} \label{sec:problemstatement}

Adversarial patches can compromise the reliability of state-of-the-art object detection. However, their impact on \ac{UAV}-related applications has been largely overlooked in the existing literature. In this section, we dive into how adversarial patch attacks can affect the reliability of the object detection task within \ac{UAV} use-case. Specifically, we first outline the specifics of our object detection task, followed by the threat model, and then assess the accuracy degradation when an adversary employs adversarial patches.

\subsection{Task} \label{subsec:task}

We consider a \ac{UAV}-based vehicle detection task, often a crucial part of aerial surveillance and tracking. We use several one-stage multilevel object detectors for this task, namely, SSD~\cite{liu2016ssd}, RetinaNet~\cite{lin2017focal}, EfficientDet~\cite{tan2020efficientdet}, and YOLOv5~\cite{ultralytics2021yolov5} in various configurations (see Tab. ~\ref{tab:problemstatement}). In consideration of the limited resources onboard and the fact that most of the aerial objects tend to be small relative to the image area, we discard the top-level feature map from the feature extraction backbones for all the object detectors and include an additional high-resolution feature map instead. Doing so reduces the size of the model and allows the model to focus on objects with a small camera footprint appropriately. In this manner, all the models that we evaluate have $\approx 7$ to $21$ million parameters, making them good candidates for deployment in \acp{UAV}. The object detectors are trained to detect vehicles belonging to four classes, namely \textit{Car}, \textit{Van}, \textit{Bus}, and \textit{Truck} on the VisDrone~\cite{visdrone} dataset, which provides images in various resolutions, lighting conditions, heights, and camera angles relative to the ground objects acquired using \ac{UAV} platforms. We use an input image size of $640\times640$ for all our experiments.

\subsection{Threat Model} \label{subsec:threatmodel}
To attack the object detectors, we assume the following threat model:

\noindent \textbf{\textit{Attacker's goal.}} The attacker's objective is to generate an adversarial patch that can be used to conceal vehicles from being identified by the object detector. The constructed patches can be attached either digitally or physically on vehicles that the attacker wishes to conceal.

\noindent \textbf{\textit{Attacker's capabilities.}} The attacker operates in a white-box setting, where they have full access to a copy of the deployed model as well as the training dataset. Realistic scenarios where these assumptions apply include a \ac{UAV} utilizing a publicly available object detection model that has been pretrained on a benchmark dataset, such as EfficientDet~\cite{EfficientDet} trained on the VisDrone dataset~\cite{visdrone}.

\subsection{The Adversarial Patch Threat}

\begin{table*}[t]
    \caption{Impact of patch attacks on \ac{UAV}-based vehicle detection task evaluated on the VisDrone test-set. Mean of $5$ runs are reported for each attack method to account for stochasticity in patch transformations.}
  \centering
  \begin{tabular}{|c|c|c|c|c|c|c|c|c|c|c|c|c|}
    \hline
    Target & Params(M) & \multicolumn{2}{c|}{Patch Free} & \multicolumn{3}{c|}{Gray Patch} & \multicolumn{3}{c|}{Random Patch} & \multicolumn{3}{c|}{Adversarial Patch}\\
    \cline{3-13}
    & & \acs{AP} & \acs{AR} & \acs{AP} & \acs{AR} & \acs{ASR} & \acs{AP} & \acs{AR} & \acs{ASR} & \acs{AP} & \acs{AR} & \acs{ASR}\\
    \hline
    Resnet50v2-SSD & $13.4$ & $0.57$ & $0.82$ & $0.20$ & $0.57$ & $0.63$ & $0.19$ & $0.56$ & $0.66$ & $0.04$ & $0.26$ & $\mathbf{0.92}$\\
    Resnet50v2-RetinaNet & $17.5$ & $0.54$ & $0.80$ & $0.24$ & $0.60$ & $0.54$ & $0.23$ & $0.62$ & $0.56$ & $0.10$ & $0.37$ & $\mathbf{0.84}$\\
    DenseNet121-RetinaNet & $14.4$ & $0.59$ & $0.82$ & $0.26$ & $0.64$ & $0.54$ & $0.26$ & $0.64$ & $0.58$ & $0.07$ & $0.32$ & $\mathbf{0.90}$\\
    EfficientDet-D3 & $12$ & $0.60$ & $0.82$ & $0.33$ & $0.69$ & $0.46$ & $0.31$ & $0.68$ & $0.52$ & $0.16$ & $0.47$ & $\mathbf{0.76}$\\
    EfficientDet-Lite4 & $15.1$ & $0.61$ & $0.83$ & $0.32$ & $0.69$ & $0.48$ & $0.31$ & $0.68$ & $0.50$ & $0.20$ & $0.54$ & $\mathbf{0.72}$\\
    YOLOv5-Small & $7$ & $0.58$ & $0.83$ & $0.29$ & $0.67$ & $0.51$ & $0.30$ & $0.68$ & $0.51$ & $0.20$ & $0.44$ & $\mathbf{0.85}$\\
    YOLOv5-Medium & $20.9$ & $0.61$ & $0.83$ & $0.30$ & $0.66$ & $0.53$ & $0.29$ & $0.67$ & $0.55$ & $0.11$ & $0.37$ & $\mathbf{0.88}$\\
    \hline
  \end{tabular}
  \label{tab:problemstatement}
\end{table*}

Our objective is to understand the impact of adversarial patch attacks on object detectors designed to work on \acp{UAV} used for vehicle surveillance. We use a $64\times64$ patch to learn adversarial information. Following an approach similar to Thys \emph{et al.}~\cite{Thys_2019_CVPR_Workshops}, we consider three losses during the patch optimization, the \ac{NPS}, \ac{TV}, and the classification score. During training, the patch is dynamically scaled for each object in a uniform range of $15$\% to $35$\% of the target bounding box area. For evaluation, we use a fixed patch area of $20$\% relative to the object bounding box.

To ensure the robustness of the attack, the following transformations are applied to the input patch during training and evaluation:
\begin{itemize}
    \item \textit{Random flip}. Horizontal and vertical
    \item \textit{Hue rotation}. Uniform range $\pm 0.08$
    \item \textit{Contrast multiplier}. Uniform range $[0.5, 1.5]$
    \item \textit{Saturation multiplier}. Uniform range $[0.5, 1.5]$
    \item \textit{Brightness adjustment}. Uniform range $\pm0.3$
    \item \textit{Per-pixel additive noise}. Uniform range $\pm0.1$
    \item \textit{Patch rotation}. Uniform range $\pm20$ degrees on camera axis
\end{itemize}

To ensure the availability of a reasonable patching area on all the objects, we preprocess the data before training by removing objects that occupy less than $0.1$\% of the original image area (i.e., area before resizing to a fixed size). 
After this prepossessing, around $90$\% of the images from the dataset are retained.

We evaluate the model \ac{AP} and \ac{AR} at \ac{IOU} threshold of $\geq0.5$ on the test dataset. Fig.~\ref{fig:prob_statement_examplepatch} shows one example of an obtained adversarial patch overlayed on the target objects.
We also evaluate the impact of adversarial patches compared to a randomly initialized patch and a gray patch as baselines, both of which employ the same size, scaling, and rotation approach as their adversarial counterparts. The objective of this comparison is to determine the extent to which the degradation in object detection accuracy is attributable to the features of the adversarial patch as compared to the occlusion caused by the addition of the patch.

We also investigate the \acf{ASR} of the generated patches.
The \ac{ASR} measures the ratio of objects successfully hidden from the object detector after adding the adversarial patch.
More specifically, it measures the ratio of correctly detected objects before and after applying the adversarial patch. 
Objects that are not correctly identified due to the added patch cause a decrease in this ratio and therefore increase \ac{ASR}. 
Similarly to the \ac{AP} metric, we take the average of \ac{ASR} at all recall thresholds for each class. We then report the \ac{ASR} as the mean of average \ac{ASR} across all classes.

Tab. \ref{tab:problemstatement} shows the object detection performance in the presence of patches. We note that the \textit{Gray} and \textit{Random} patch baselines perform similarly, registering on average, $53$\% and $55$\% \ac{ASR} respectively, even with a patchable area of only $20$\% per object. This points to the fragility of \ac{UAV}-based object detection and shows that ensuring robustness is in fact a challenging task in this scenario.
It is possible to note that the adversarial patch severely impacts the accuracy of all object detection models considered in our work, achieving on average $84$\% \ac{ASR} for all models considered in this paper, an increase of over $1.5\times$ the baseline results. In fact, as Tab.~\ref{tab:problemstatement} shows, the \ac{ASR} approaches $90$\% for some models.

\begin{figure}[t!]
    \centering
    \begin{subfigure}[b]{0.505\linewidth}
        \centering
        \includegraphics[width=\textwidth]{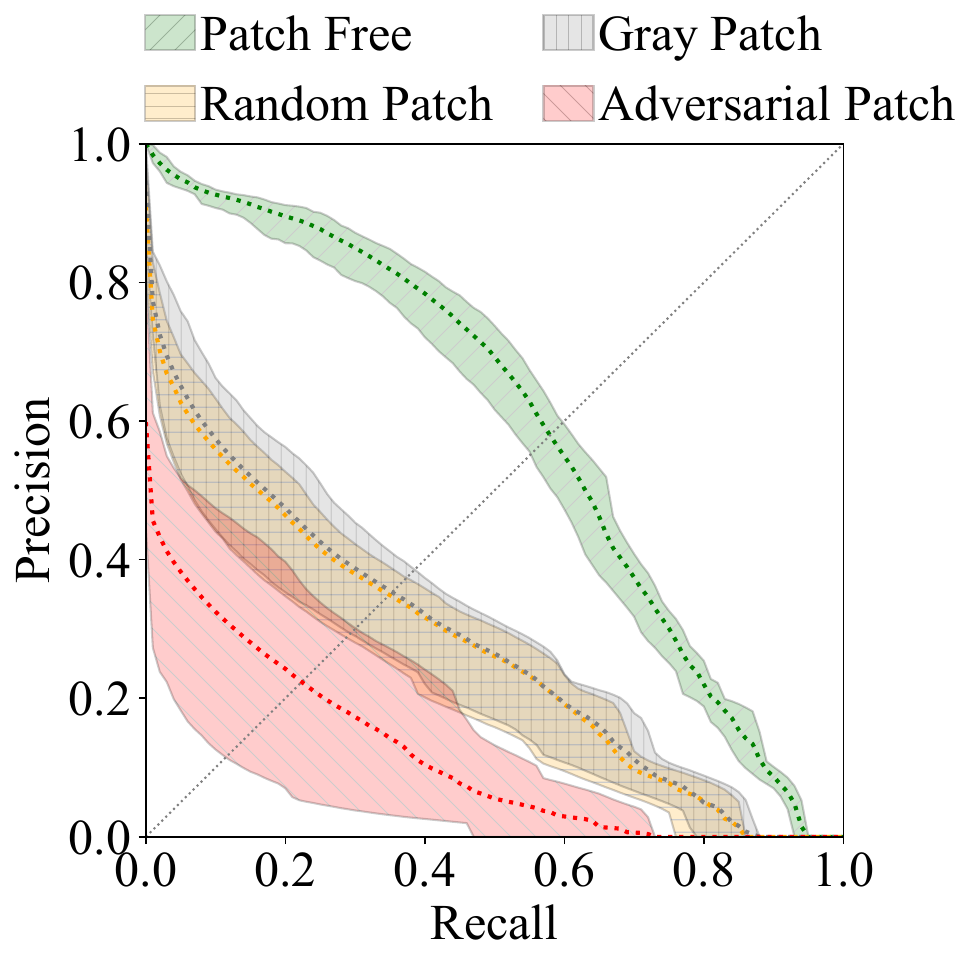}
        \caption{Averaged Precision \textit{vs.} Recall}
        \label{fig:probstatement_vehicle_prec_recall}
    \end{subfigure}
    \begin{subfigure}[b]{0.475\linewidth}
        \centering
        \includegraphics[width=\textwidth]{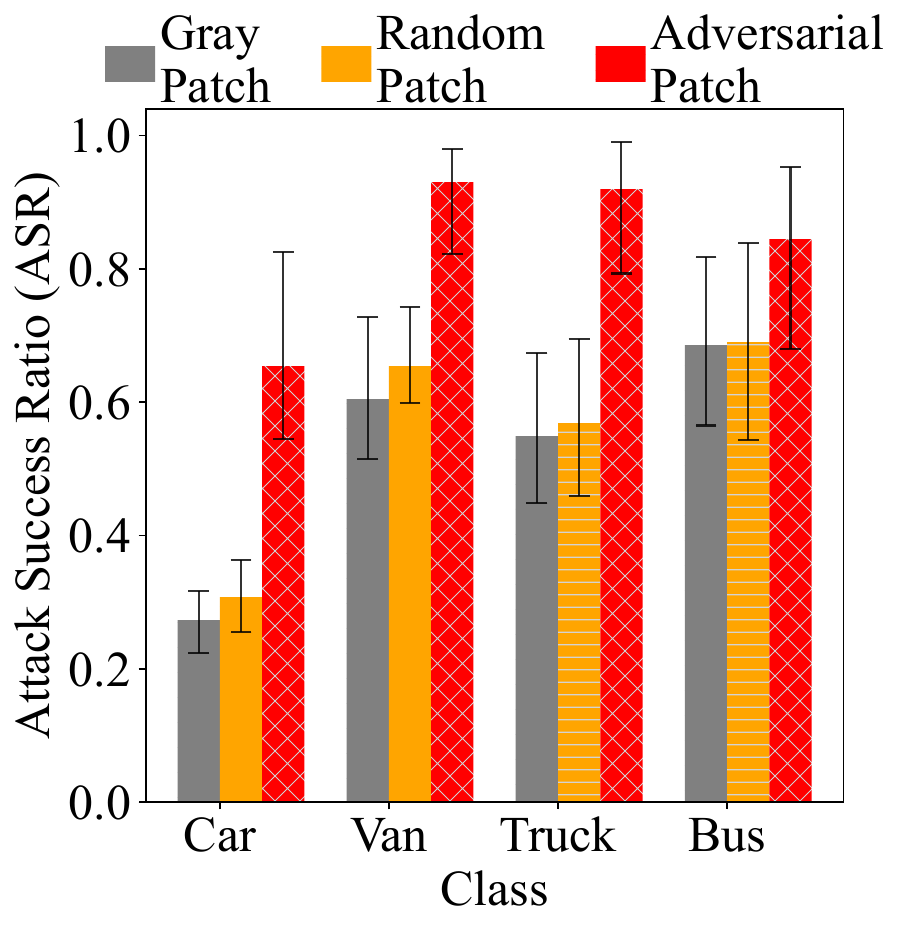}
        \caption{Mean \acs{ASR} by Class}
        \label{fig:probstatement_vehicle_asr_class}
    \end{subfigure}
    \caption{Aggregated results across all models showing the impact of patch attacks on vehicle detection performance on the VisDrone test set. Each attack method was evaluated $5$ times per model. Shaded regions and the error bars denote minimum and maximum values across all models for the respective attack method.}
    \label{fig:probstatement_vehicle}
\end{figure}

Fig.~\ref{fig:probstatement_vehicle_prec_recall} shows the consolidated impact of patch attacks on the accuracy of all object detectors considered in our evaluation. Notably, adversarial patches affect both \textit{Precision} and \textit{Recall} metrics catastrophically in our use case. In Fig.~\ref{fig:probstatement_vehicle_asr_class}, it can be observed that the \textit{Car} class is least impacted by the baseline occlusion patches and the contrast between the effect of adversarial and non-adversarial patches is the strongest there. We argue that this is because the VisDrone dataset contains significantly more instances of the \textit{Car} class, compared to others. This biases the detector to identify the majority class more effectively, reducing the efficacy of occlusion attacks. Similarly, the adversarial patch is biased to be more effective in concealing the instances of the \textit{Car} class during training, since the patch is applied to them more frequently during training.

\begin{figure*}[t!]
\centering
\includegraphics[width=1.0\textwidth]{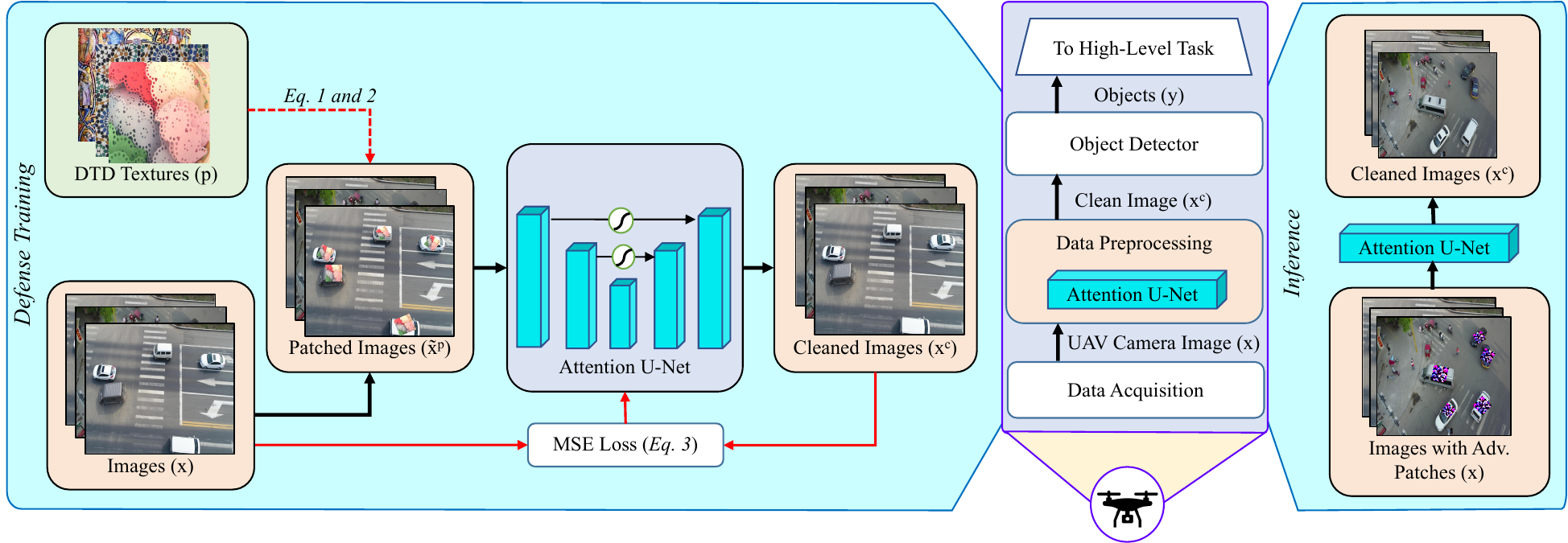}
\caption{Proposed model-agnostic mechanism for adversarial patch defense on \ac{UAV} object detectors. 
The Attention U-Net model is trained to reconstruct regions masked by \acs{DTD} patches in the input image. The inference phase uses the built model to preprocess the input image and restore the object regions altered by an adversary to the estimated object pixels without the patch.}
\label{fig:proposal}
\end{figure*} 

\subsection{Discussion}
In this section, we have evaluated the impact of patch attacks on commonly used object detectors in the context of a \ac{UAV}-based vehicle detection task. Our evaluation has yielded two key findings that underscore the need for a defense mechanism to counteract the threat posed by patches. First, our experiments have shown that adversarial patches can significantly impair the performance of \ac{UAV} object detection, posing a considerable threat to upstream tasks. Second, we also observe a notably sharp decline in performance even in the presence of random or gray patch occlusions, although not as severe as with adversarial patches. These findings suggest that an external defense mechanism may be necessary to ensure reliability in \ac{UAV} object detection pipelines when internal protection, such as an adversarially trained model, is unavailable.
Object detection plays a critical role in enabling automation on \ac{UAV} applications. Therefore, the provision of a reliable and resilient object detection procedure that is feasible to implement on \acp{UAV} is a must to ensure their reliability.
\section{A Defense Mechanism Against Adversarial Patch Attacks} \label{sec:proposal}
Current adversarial patch defense approaches in the literature generally focus on defending against a specific patch generation approach, lacking robustness to be applied in real-world conditions.
In such a case, the attacker can generate new adversarial patches as needed to evade detection. Consequently, these solutions must be updated as new patch generation approaches arise or when the downstream model is changed.

Our proposal aims to defend \ac{UAV} object detection against adversarial patch attacks that aim to conceal an object from the detector. We know that a significant portion of the adversarial patch threat on object detection performance comes from the presence of adversarial patterns on the patch. However, as we have shown in Sec.~\ref{sec:problemstatement}, even non-adversarial patches can potentially impact the performance of an object detector due to their occlusion effect alone. The impact due to occlusion is expected to be severe for \ac{UAV} based object detection since the profile of objects visible to the \ac{UAV} camera tends to be very small during missions. It motivates us towards a defense approach that is capable of mitigating both these aspects.

We rethink adversarial patch defense from the viewpoint of image restoration. We aim to achieve this in a model-agnostic manner. To that effect, our approach does not require prior access to the details of the adversarial attack, such as adversarial patterns that might be specific to an object detector. Our only assumption is that the attacker uses patches that are positioned on the objects of interest, occluding them partially. Our objective is to recover the object pixels occluded by the patch and in doing so, mitigate the effect of the patch on the object detector. In this manner, we aim to abstract away the details on the patch itself, modeling our defense approach as an occluded object reconstruction problem. Doing so helps us in two ways. First, our defense mechanism is able to defend against adversarial and non-adversarial patch occlusions on objects of interest. Second, our approach allows us to maintain a model-agnostic nature, and once deployed does not require to be updated in response to changes in the object detection pipeline. Additionally, our solution is simple and lightweight in order to form the preprocessing stage in the real-time object detection pipeline onboard \acp{UAV}.

The implementation of our proposed scheme is shown in Fig.~\ref{fig:proposal}. We use the Attention-UNet~\cite{oktay2018attention} autoencoder architecture for reconstructing the input image in the preprocessing stage. During training, we encourage the model to identify and ignore the patch pixels and reconstruct the image with the object pixels instead. In the inference pipeline, our preprocessing stage recovers the object pixels occluded by patches in the input image before sending it to the downstream object detection model.

\subsection{Training}
\label{sec:trainingdefense}
\begin{figure}[t!]
    \centering
    \begin{subfigure}[b]{0.45\columnwidth}   
        \centering 
        \includegraphics[width=0.99\textwidth]{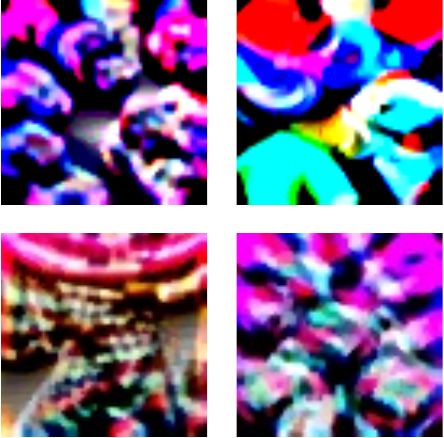}
        \caption{Adversarial patches}
        \label{fig:example_with_vehicle_patch}
    \end{subfigure}
    \hfill
    \begin{subfigure}[b]{0.45\columnwidth}   
        \centering 
        \includegraphics[width=0.98\textwidth]{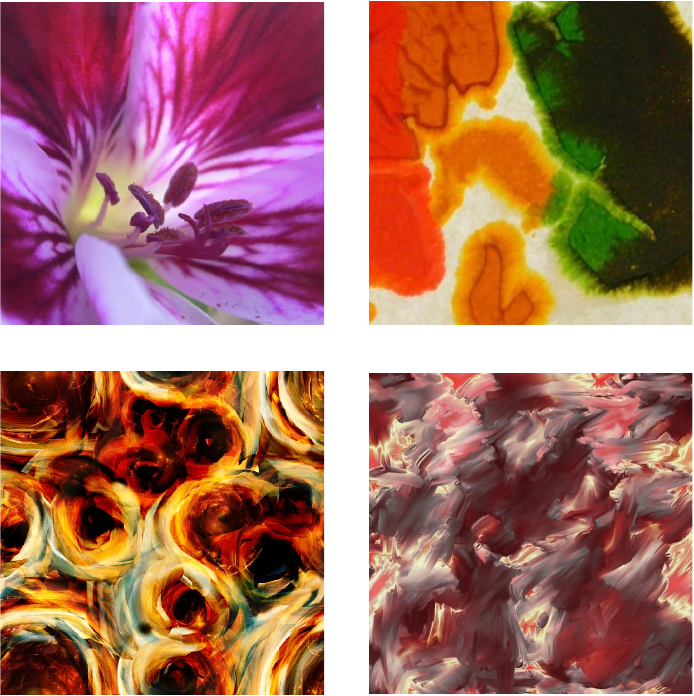}
        \caption{\ac{DTD} textures}
        \label{fig:example_with_vehicle_patch_removed}
    \end{subfigure}
    \caption{Example images from the \acf{DTD} (right) with textural similarities to adversarial patches (left).}
    \label{fig:dtdsample}
\vspace{-3mm}
\end{figure}

We train our autoencoder on the VisDrone dataset for the vehicle restoration task. To occlude vehicles, we dynamically add patches to ground-truth bounding boxes following a similar approach and set of transformations described in Sec.~\ref{sec:problemstatement}. For our task, we desire to be independent of the adversarial patterns on a patch yet be able to detect and remove them. To that end, we realize that adversarial patterns are typically comparable to textures. Therefore, we use the \ac{DTD} as a source of textures that are applied to objects as patches during training of our defensive scheme. Fig.~\ref{fig:dtdsample} shows an example of images from the \ac{DTD} dataset bearing textural similarities to the adversarial patches learned in Sec.~\ref{sec:problemstatement}.

At training time, let $x$ be an image from a given training batch $D$ 
such that $x \in [-1, 1]^{H\times W\times 3}$ where $H$ and $W$ denote the height and width of the image.
We generate the autoencoder input (Fig.~\ref{fig:proposal}, \textit{Patched Images ($\tilde{x}^{p}$)}) based on the following functions:

\begin{equation}
     \begin{aligned}
        \tilde{p_y}, s_y, l_y = T(p, y)
    \end{aligned}
    \label{eq:patchtranform}
\end{equation}
\begin{equation}
     \begin{aligned}
        \tilde{x}^{p} = A(\tilde{p_y}, s_y, l_y, x, y)
    \end{aligned}
    \label{eq:patchedimage}
\end{equation}

\noindent where $T(p, y)$ is a set of stochastic transformations that are applied on a texture patch $p$ obtained from the \ac{DTD} dataset for each object $y \in x$. Subsequently, a patch application function $A(\tilde{p_y}, s_y, l_y, x, y)$ applies the transformed patch $\tilde{p_y} \in \mathbb{R}^{s_y\times s_y\times 3}$ at location $l_y$ for each object $y$, where $s_y$ is the size.
The location $l_y$ is placed within the bounding box $b_y$ of the object $y$, such that $b_y > s_y$.

The objective is to restore a given patched image $\tilde{x}^{p}$ to its unpatched counterpart $x$ (see Fig.~\ref{fig:proposal}). We use the pixel-wise Mean Squared Error (MSE) loss
\begin{equation}
     \begin{aligned}
        \mathcal{L}(x, x^{c}) \propto \sum_{i}^{H}\sum_{j}^{W}(x_{ij}-x^{c}_{ij})^{2}
    \end{aligned}
    \label{eq:loss}
\end{equation}
\noindent where $x^c$ is the reconstructed output of the autoencoder.

\subsection{Defense Implementation for UAVs}
It is important to keep the defensive model lightweight to reduce the computational cost incurred due to the addition of a preprocessing stage while maintaining the feasibility of the solution to be deployed in a real-time object detection pipeline onboard a \ac{UAV}. To that extent, we modify the standard model architecture.
We use an EfficientNet-B0~\cite{tan2019efficientnet} backbone pre-trained on the ImageNet benchmark as the encoder for the Attention-UNet. Taking into account the small area footprint of ground objects relative to the \ac{UAV} camera, we further discard the top-level feature map from the backbone and instead include a high-resolution map from the prior layers to enable the availability of small object details to the decoder. Doing so also reduces the number of parameters in the encoder layers. We use a slim decoder to further reduce the inference time, with only one decoder convolution per level.  We use multiplicative attention in the attention module. We do not use any pooling layers and use the Hard-Swish~\cite{howard2019searching} as an activation function in the decoder for faster computation. 
Our decoder has a $5$-level configuration with $16$, $32$, $64$, $128$, and $256$ filters, respectively. As a result, our model has only $\mathbf{\approx 1.2}$ \textbf{million} parameters, making it suitable for deployment in \acp{UAV}.

We do not freeze the backbone during training; the complete architecture is trained on the VisDrone dataset.
The model training was executed for $200$ epochs with a batch size of $16$ images. SGD with momentum was used for optimization, with a cosine annealed learning rate schedule.

\section{Evaluation} \label{sec:evaluation}
In the context of the \ac{UAV} vehicle detection task, our evaluation aims to answer the following \acp{RQ} in this section:

\begin{itemize}
    \item (\textbf{\ac{RQ}1}) \textit{How well does the proposed defense scheme perform?}
    \item (\textbf{\ac{RQ}2}) \textit{What are the processing costs of our proposal?}
    \item (\textbf{\ac{RQ}3}) \textit{How well does our approach perform in the physical domain?}
\end{itemize}

\subsection{Defending Against Adversarial Patches}
\label{sec:defenseeval}
To answer \ac{RQ}1, we evaluate object detection performance when using the cleaned images generated by the autoencoder.

\begin{figure}[t!]
\centering
\includegraphics[width=0.99\columnwidth]{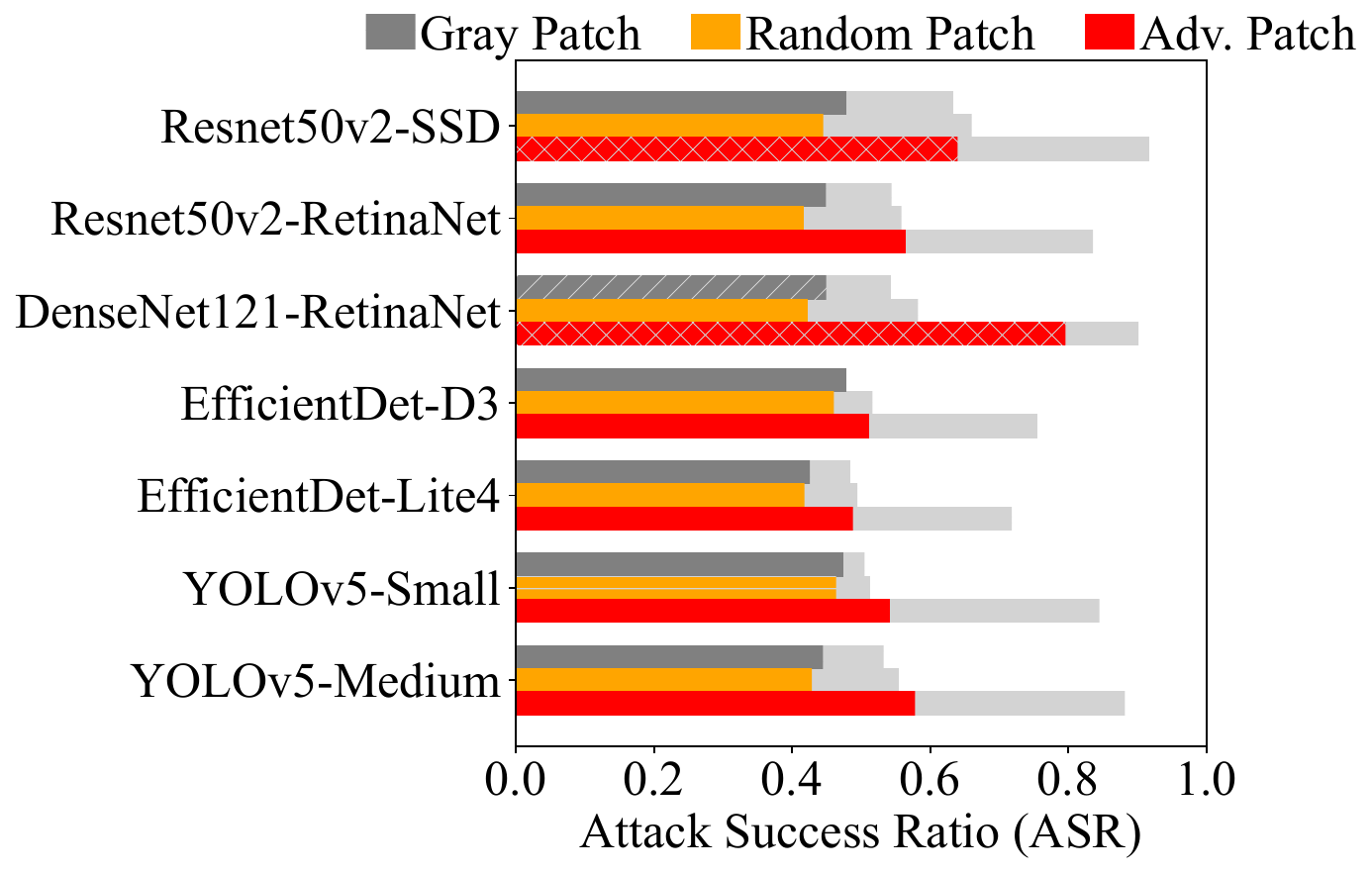}
\caption{Performance of the proposed scheme evaluated against individual object detectors, as compared to the performance without defense shown here as a light-gray shadow. Mean of $5$ runs are reported for each attack method to account for stochasticity in patch transformations.}
\label{fig:proposalASRallschemes}
\end{figure}

Fig.~\ref{fig:proposalASRallschemes} shows that the proposed defense approach can, in practice, reduce adversarial and non-adversarial patch occlusions in the input image at the preprocessing stage, resulting in previously hidden objects being detected by the downstream object detector, reducing the \ac{ASR} of an adversary.
On average, our method reduces the \ac{ASR} of adversarial patch attacks in all models from $84$\% to $59$\%, a relative reduction of $\approx 30$\%.
As a result, our model significantly reduces the impact of adversarial patch attacks on the reliability of \ac{UAV} object detection.
Qualitatively, a translucent effect can be seen in place of the patch due to object pixels restored by our approach (see Fig.~\ref{fig:prob_statement_examplepatch}).
For non-adversarial patches, our method registered a $\approx 13$\% and $\approx 21$\% relative improvement on \ac{ASR} averaged across all models for \textit{Gray} and \textit{Random} patch attack baselines, respectively.

\begin{figure}[t!]
    \centering
    \begin{subfigure}[b]{0.505\columnwidth}   
        \centering 
        \includegraphics[width=0.99\textwidth]{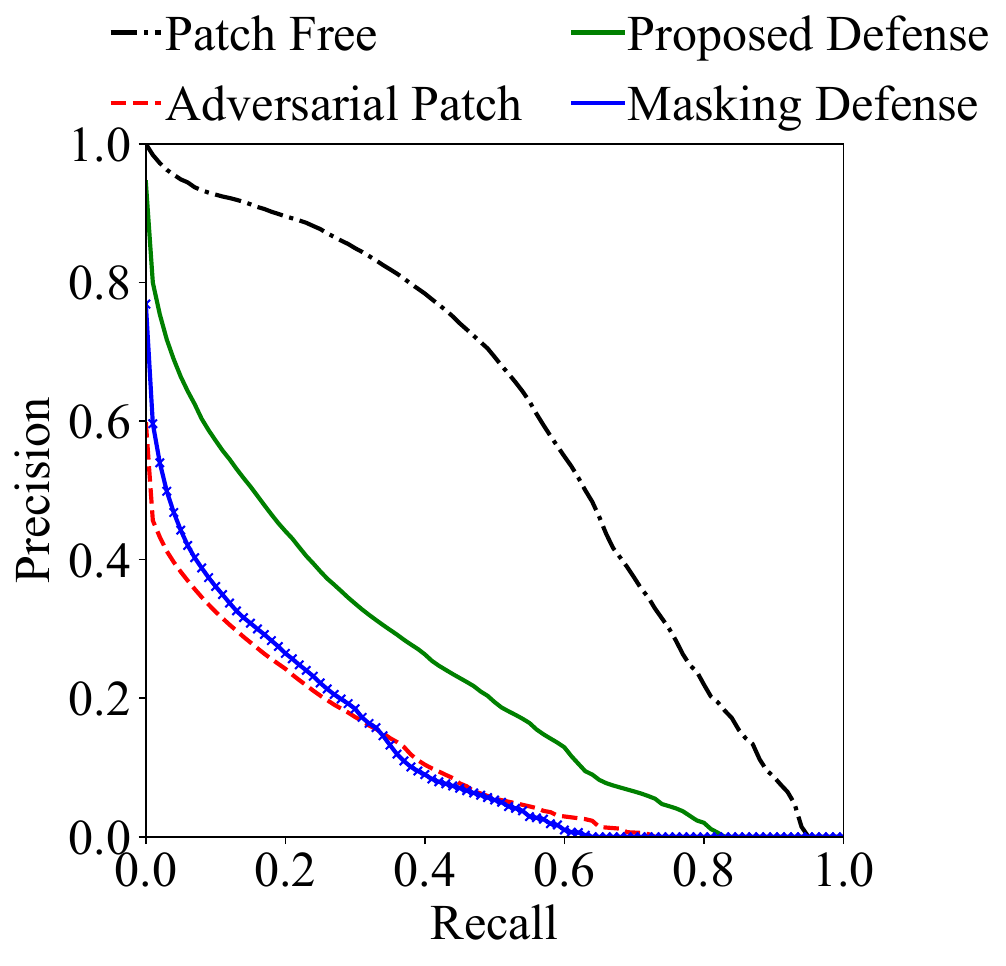}
        \caption{Averaged Precision \textit{vs.} Recall}
        \label{fig:proposalASR_car}
    \end{subfigure}
    \begin{subfigure}[b]{0.475\columnwidth}   
        \centering 
        \includegraphics[width=0.99\textwidth]{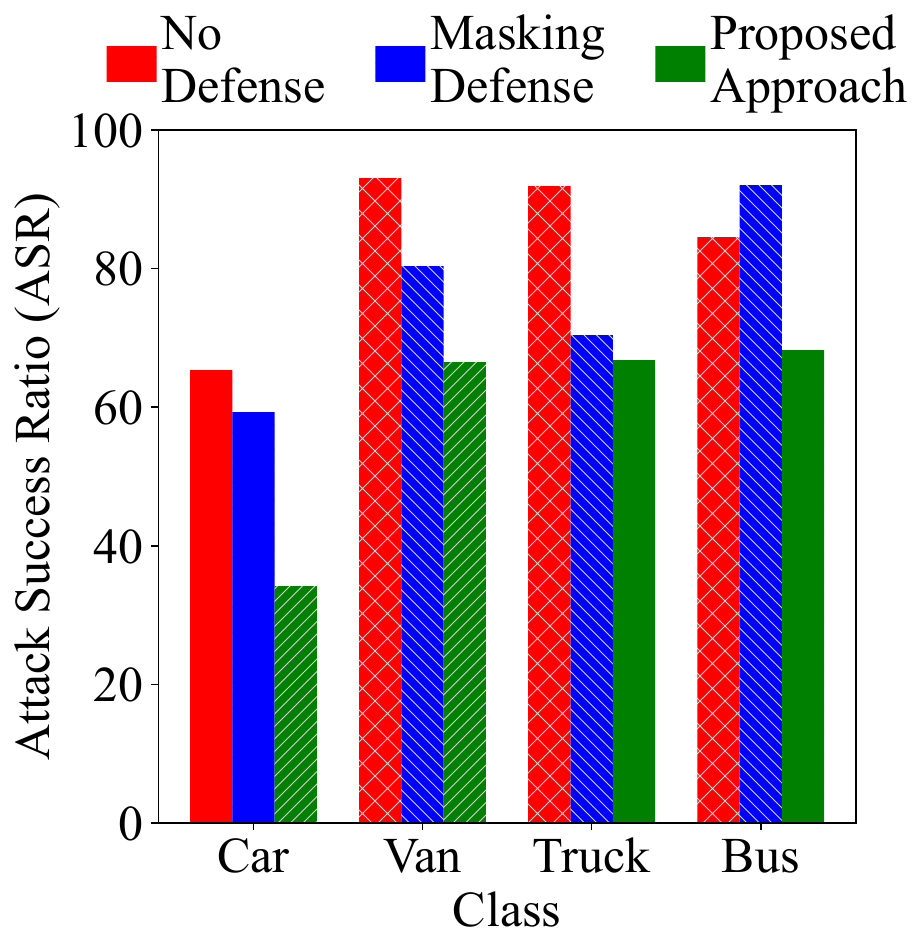}
        \caption{Mean \acs{ASR} by class}
        \label{fig:proposalASR_vehicle}
    \end{subfigure}
    \caption{The proposed method in comparison to a model-agnostic patch-masking approach based on segmentation.}
    \label{fig:proposalmaskASR}
\end{figure}

\noindent \textbf{\textit{Comparision:}} Our solution is closely related to \textit{Patch Masking} defense techniques similar to~\cite{Liu_2022_CVPR, chiang2021adversarial}, in which the identified patch is masked out from the input image rather than restoring it. For a fair comparison with a pixel-masking-based solution, we keep our approach model-agnostic by avoiding knowledge about the downstream object detector in contrast to~\cite{Liu_2022_CVPR, chiang2021adversarial}. We construct a single-stage pixel masking approach, using Attention-UNet as before, but for a patch segmentation task and train it in an identical manner. The obtained segmentation mask is inverted and multiplied pixel-wise with the input image in the preprocessing stage.
Fig.~\ref{fig:proposalmaskASR} shows the average \acf{ASR} impact on all object detectors in the presence of adversarial patches. It is evident that a standalone model-agnostic approach based on segmentation-based masking alone is insufficient for defense and requires additional measures or assumptions to be effective. On the other hand, the proposed defense solution can defend the \ac{UAV} object detectors in a model-agnostic manner without having prior access to the adversarial patches or object detectors. Having a tiny single-stage defense mechanism also helps in deployment onboard \acp{UAV}.

To evaluate the processing costs of our adversarial patch defense solution (\ac{RQ}2), we compare the average execution time of the object detection pipeline with and without defense. Without any quantization or optimization, our solution incurred an average additional processing time of only $\approx 4$\% per image during inference on the VisDrone test set.
Consequently, we are able to provide reliable object detection for \acp{UAV} without incurring high additional processing costs.

\subsection{Physical Domain Experiments}
\label{sec:physical}
For the physical domain scenario (\ac{RQ}3), we constructed a controlled environment laboratory setup wherein we used toy model vehicles ($1$:$50$ downscaling ratio) and imaged them from various angles at heights ranging from $1$ to $6$ feet using a standard high-definition camera. In practice, our laboratory setup closely mimics the aerial views of vehicles taken from a \ac{UAV}-based camera such that the images of the vehicles achieve the same camera footprint as observed in the VisDrone dataset. Our setup also allows us to quickly print patches on a standard high-definition printer and place them directly on the toy models, instead of using actual cars which requires a much larger and complex printing setup and overlaying process.
\begin{figure}[t!]
    \centering
    \begin{subfigure}[b]{0.49\columnwidth}   
        \centering 
        \includegraphics[width=0.99\textwidth,trim={1cm 2cm 1cm 1cm},clip]{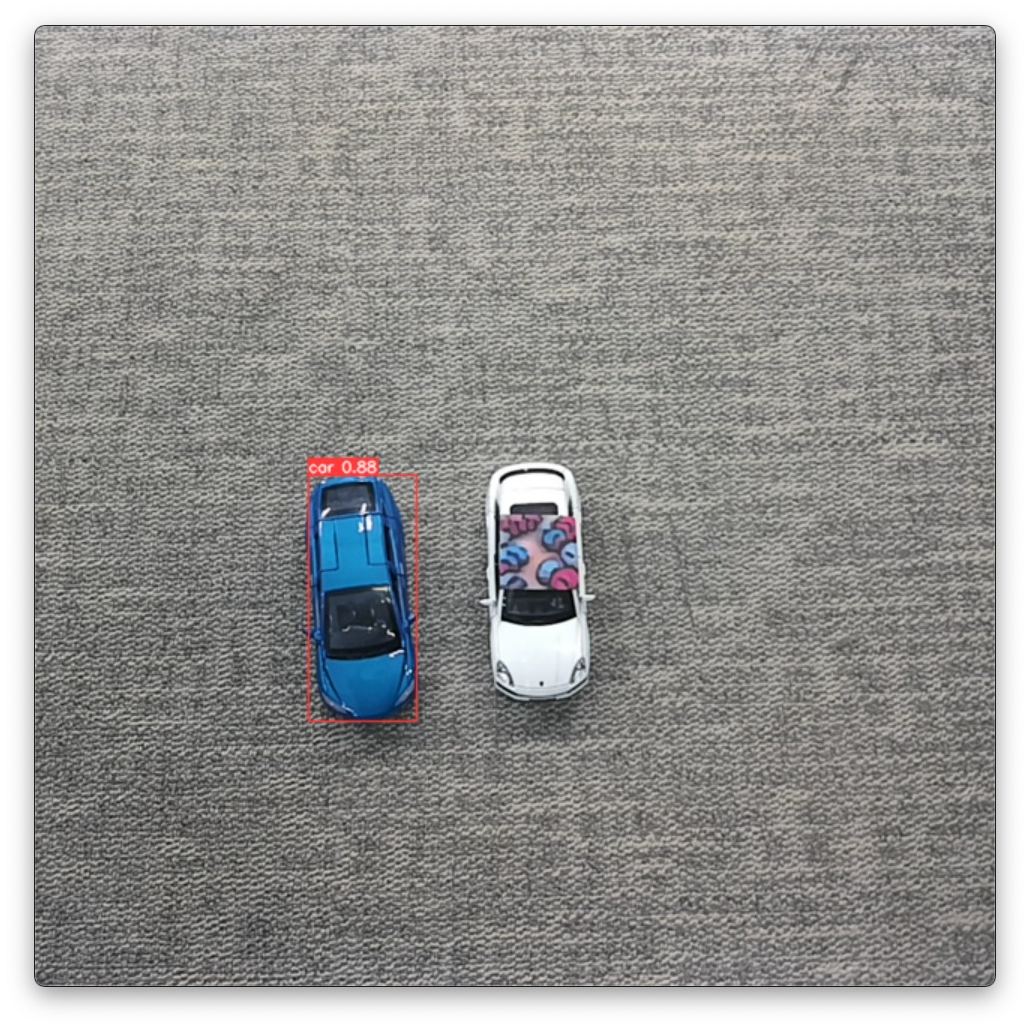}
        \caption{Without defense}
        \label{fig:physicalASR_car_line}
    \end{subfigure}
    \begin{subfigure}[b]{0.49\columnwidth}   
        \centering 
        \includegraphics[width=0.99\textwidth,trim={1cm 2cm 1cm 1cm},clip]{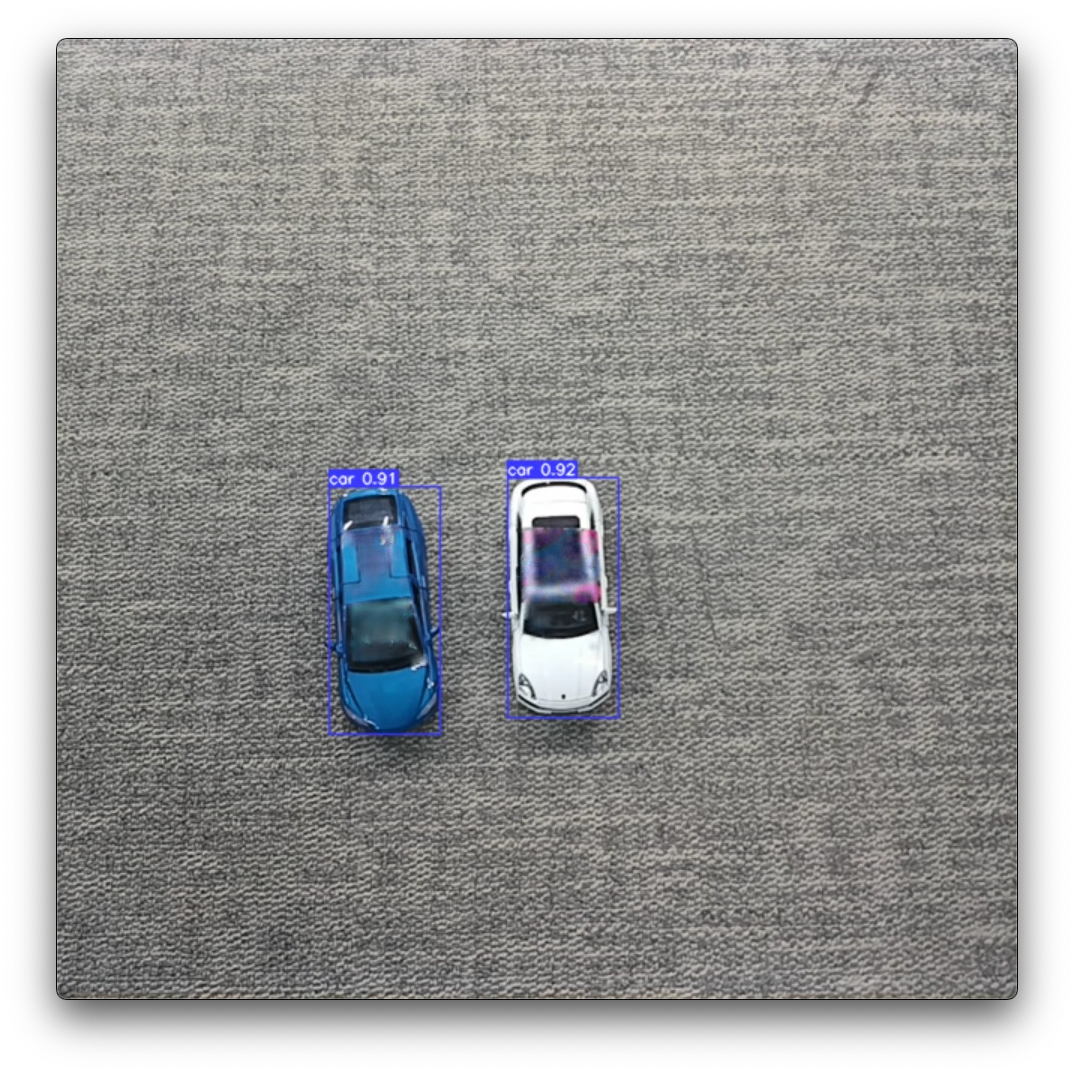}
        \caption{With defense}
        \label{fig:physicalexp_post}
    \end{subfigure}
    \caption{Example of a physically printed patch (left) affecting the accuracy of a YOLOv5s object detector. Our defense approach can identify and remove the physical patch (right) from the camera image, making the vehicle detectable.}
    \label{fig:physicalexp}
\end{figure}
A YOLOv5s model was trained on the collected toy dataset comprising around a hundred images collected for the task. Considering the limited number of images, a VisDrone checkpoint was used for finetuning, and data augmentations were heavily employed. Similarly, we trained adversarial patches to hide the cars. Fig.~\ref{fig:physicalexp} shows the performance of our approach on an image from the hold-out validation set with a physically printed adversarial patch. As can be seen, the physical patch successfully conceals the car from the object detector, while the proposed solution can digitally remove the patch, causing the object to be detected. This scaled-down laboratory experiment serves as a proof of concept that the effectiveness of our approach extends to the physical domain in the case of \ac{UAV}-based object detection.

\section{Conclusion} \label{sec:conclusion}

Object detection is indispensable for \ac{UAV} applications.
Therefore, deployed solutions must be resilient and reliable against adversaries.
In this paper, we have shown that adversarial patches significantly degrade the accuracy of \ac{UAV} object detection, posing a significant threat to their complete automation.
To address this challenge, we have proposed a defense model that formulates adversarial patches as an image restoration task.

The proposed scheme is implemented without using adversarial information from the object detectors during the training phase and is executed as a preprocessing task during the inference phase. This makes our method effective on multiple object detectors as shown in the paper. The evaluations carried out have shown the efficacy of the approach, significantly decreasing the success of attacks without incurring a significant impact on processing costs. Notably, the effectiveness of our method is determined by the richness and variety of textural patterns the defensive model is exposed to during training. Adversarial patterns that do not follow the textures understood by the model are likely to escape the defense mechanism.

In future work, we would investigate our approach in additional tasks. A key insight of our method comes from using a relatively simple but lightweight image restoration technique in the form of pixel-wise image reconstruction. It would be beneficial to employ more advanced techniques from the image inpainting literature to improve the effectiveness of this method. A single-stage approach that combines masking to detect and abstract away the adversarial patterns followed by inpainting is an interesting direction to explore in the future.

\small
\bibliographystyle{IEEEtran}
\bibliography{IEEEabrv, references}

\end{document}